\def\year{2019}\relax
\documentclass[letterpaper]{article} 
\usepackage{aaai19}  
\usepackage{times}  
\usepackage{helvet}  
\usepackage{courier}  
\usepackage{url}  
\usepackage{graphicx}  
\usepackage{pifont}
\usepackage{amsmath}
\usepackage{amssymb}
\usepackage{subfig}
\usepackage{makecell}
\newcommand{\xmark}{\ding{55}}
\title{\emph{GestARLite}: An On-Device Pointing Finger Based Gestural Interface for Smartphones and Video See-Through Head-Mounts}
\author{}
\author{Varun Jain, Gaurav Garg, Ramakrishna Perla and Ramya Hebbalaguppe\\
TCS Research, India \\
\texttt{\{varun.in, ga.gaurav, r.perla, ramya.hebbalaguppe\}@tcs.com}
}

\begin{document}
\maketitle
\begin{abstract}
    Hand gestures form an intuitive means of interaction in Mixed Reality (MR) applications. However, accurate gesture recognition can be achieved only through state-of-the-art deep learning models or with the use of expensive sensors. Despite the robustness of these deep learning models, they are generally computationally expensive and obtaining real-time performance on-device is still a challenge. To this end, we propose a novel lightweight hand gesture recognition framework that works in First Person View for wearable devices. The models are trained on a GPU machine and ported on an Android smartphone for its use with frugal wearable devices such as the Google Cardboard and VR Box. The proposed hand gesture recognition framework is driven by a cascade of state-of-the-art deep learning models: MobileNetV2 for hand localisation, our custom fingertip regression architecture followed by a Bi-LSTM model for gesture classification. We extensively evaluate the framework on our \emph{EgoGestAR} dataset. The overall framework works in real-time on mobile devices and achieves a classification accuracy of $80\%$ on EgoGestAR video dataset with an average latency of only $0.12$ \textit{s}.
\end{abstract}

\section{Introduction}
Over the past few decades, information technology has transitioned from desktop to mobile computing. Smartphones, tablets, smart watches and Head Mounted Devices (HMDs) are slowly replacing the desktop based computing. There has been a clear shift in terms of computing from office and home-office environments to an anytime-anywhere activity~\cite{schmalstieg2016augmented}. Mobile phones form a huge part of lives: the percentage of traffic on the internet generated from them is overtaking its desktop counterparts~\footnote{https://stonetemple.com/mobile-vs-desktop-usage-study/}. Naturally, with this transition, the way we interact with these devices also has evolved from keyboard/mice to gestures, speech and brain computer interfaces. In a noisy outdoor setup, speech interfaces tend to be less accurate, and as a result the combination of hand gestural interface and speech are of interest to most HCI researchers. Hand gesture recognition on a real-time feed or a video is a form of activity recognition, and we specifically target models that can work on smartphones.

\begin{figure}[ht]
    \centering
    \includegraphics[width=0.9\linewidth]{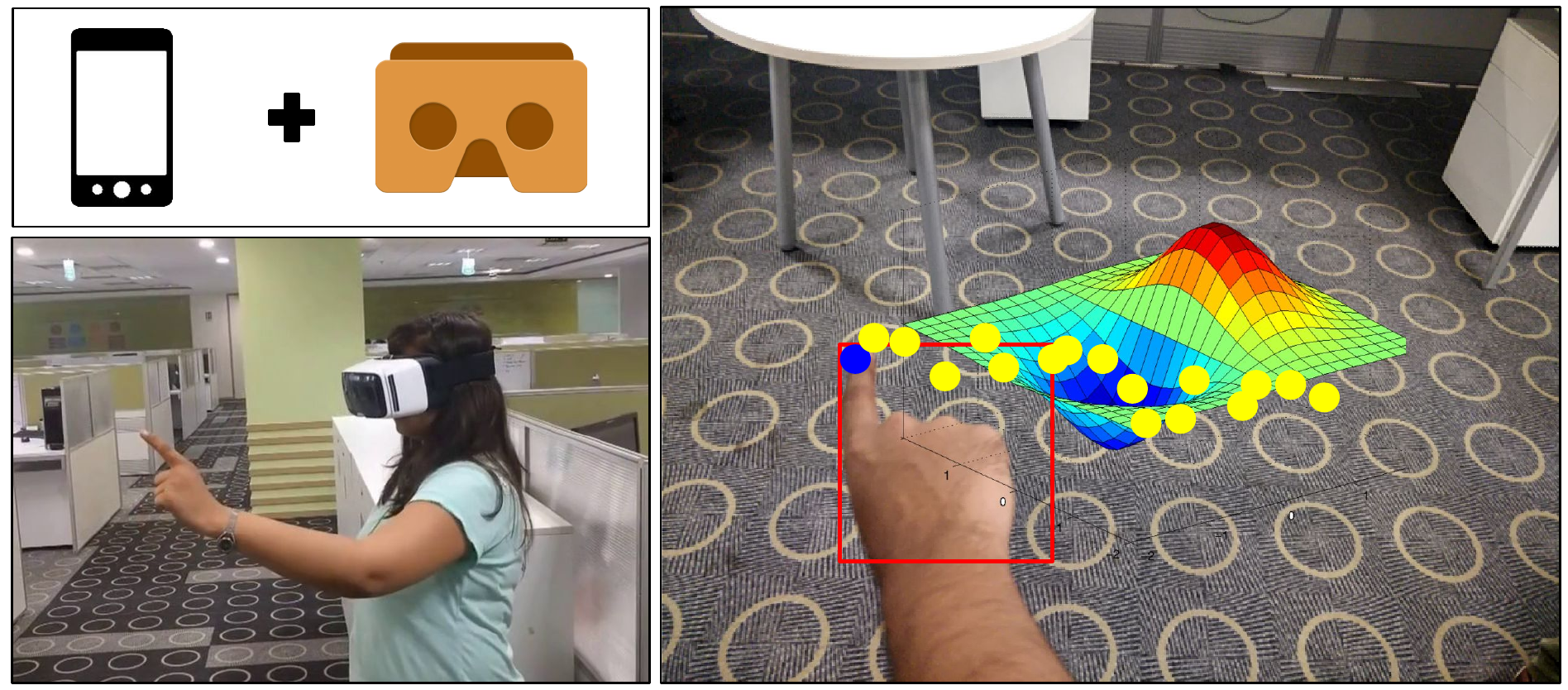}
    \caption{\emph{GestARLite}: \textit{(left)} Our framework works on commercial off-the-shelf smartphones attached to generic video see-through head-mounts like the Google Cardboard. \textit{(right)} Illustration of our MR application on Google Cardboard for Android devices: The application shows enhanced visualisation of data in the form of 3D graphs. Manipulation of the graph is done through pointing hand gestures like a left swipe as visualised by the gesture trail (in yellow).}
    \label{fig:introfigure}
\end{figure}

\begin{figure*}[ht]
    \centering
    \includegraphics[width=0.85\linewidth]{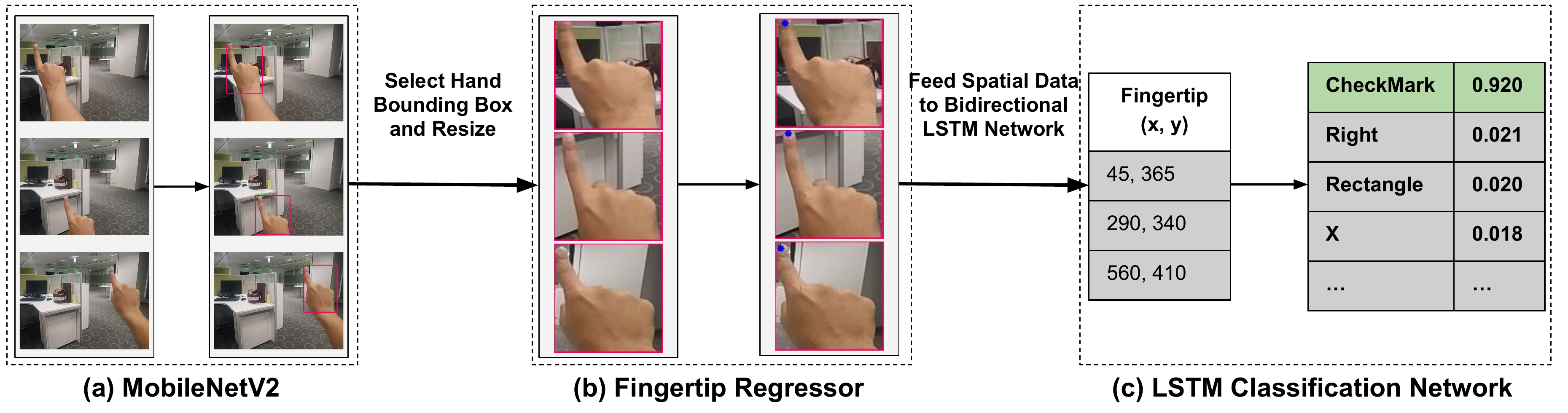}
    \caption{\emph{GestARLite}, our proposed pointing gestural framework consists of: \textit{(a)} The \textit{MobileNetV2} detector~\cite{sandler2018mobilenetv2} block that takes RGB image as an input and outputs a hand candidate bounding box, \textit{(b)} we crop and resize the hand candidate and feed it to our Fingertip Regressor architecture (refer Figure~\ref{fig:regressionframework}) for accurately localising the fingertip, and, \textit{(c)} a Bi-LSTM network for the classification of fingertip detections on subsequent frames into different gestures.}
    \label{fig:framework}
\end{figure*}

Expensive AR/MR devices such as the Microsoft HoloLens, Daqri and Meta Glasses provide a rich user interface by using recent hardware advancements. They are equipped with a variety of on-board sensors including multiple cameras, a depth sensor and proprietary processors. This makes them expensive and unaffordable for mass adoption.

In order to provide a user friendly interface via hand gestures, detecting hands in the user's Field of View (FoV), localising certain keypoints on the hand, and understanding their motion pattern has been of importance to the vision community in recent times. Despite having robust deep learning models to solve such problems using state-of-the-art object detectors and sequence tracking methodologies, obtaining real-time performance on-device is still a challenge owing to resource constraints on memory and processing. 

In this paper, we propose a computationally effective hand gesture recognition framework that works without depth information and the need of specialised hardware, thereby providing mass accessibility of gestural interfaces to the most affordable video see-through HMDs. These devices provide VR/MR experiences by using stereo rendering of the smartphone camera feed but have limited user interaction capabilities~\cite{hegde2016gestar}. 

Industrial inspection and repair, tele-presence, and data visualisation are some of the immediate applications for our framework and we aim to design a mobile-based hand gesture recognition framework which can work in real-time and has the benefit of being able to work in remote environments without the need of internet connectivity. To demonstrate the generic nature of our framework, we evaluate the detection of $10$ complex gestures performed using the pointing hand pose with a sample Android application. Figure~\ref{fig:introfigure} shows a real world application of 3D data visualisation on an Android smartphone to be used with Google Cardboard device.

The summary of key contributions is as follows:
\begin{enumerate}
    \item We propose \emph{GestARLite}: an on-device gestural interface based on fingertip regression via a cascade of neural network blocks, consisting of, MobileNetV2 for hand detection, our architecture for fingertip regression followed by a Bi-LSTM for gesture classification. This approach is marker-less and uses only RGB data without depth information. The trained models are ported on to an Android device for validating the accuracy and real-time performance of the proposed framework.
    \item EgoGestAR: a dataset of spatio-temporal sequences representing $10$ gestures suitable for MR applications. View our demo and the dataset at: \url{https://varunj.github.io/gestarlite}
\end{enumerate}

\begin{figure*}[ht]
    \centering
    \includegraphics[width=0.8\linewidth]{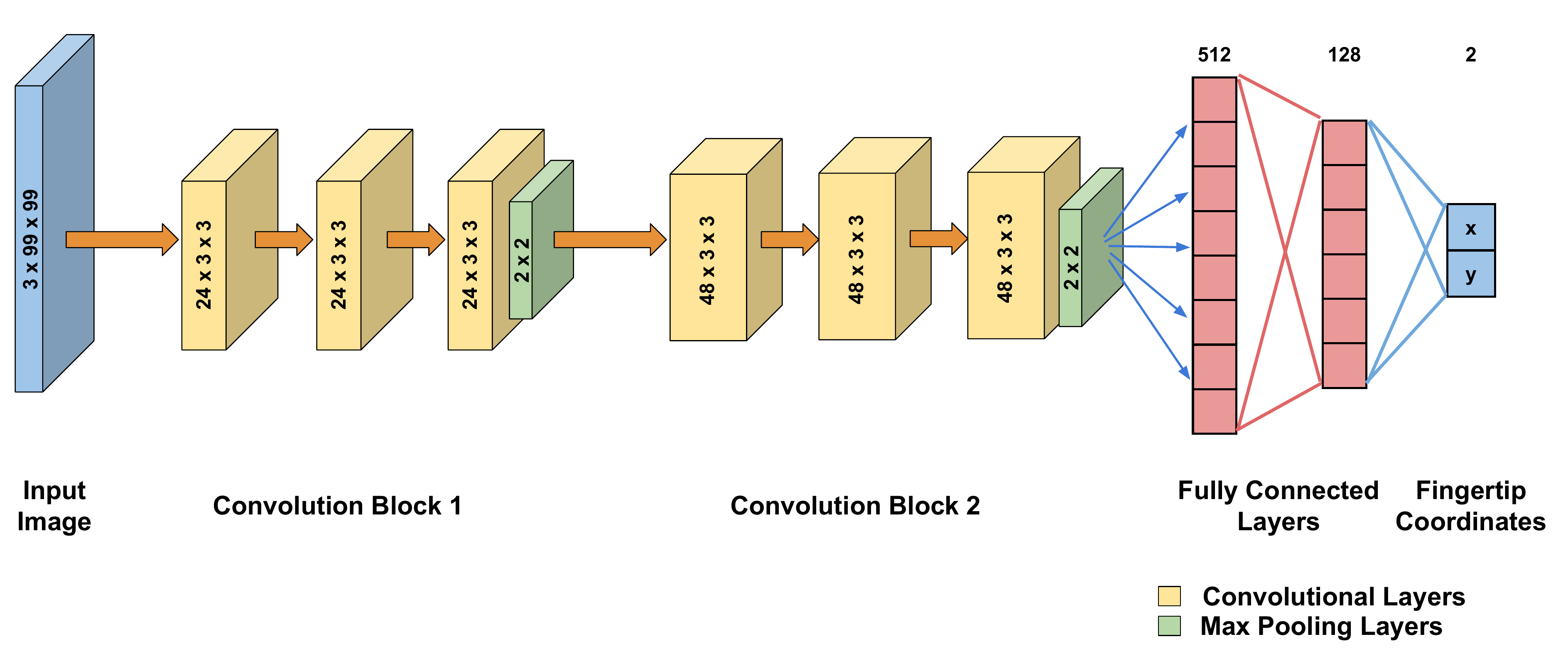}
    \caption{Overview of our proposed fingertip regressor architecture for fingertip localisation. The input to our network are $3\times99\times99$ sized RGB images. Each of the $2$ convolutional blocks have $3$ convolutional layers each followed by a max-pooling layer. The $3$ fully connected layers regress over fingertip spatial location.} 
    \label{fig:regressionframework}
\end{figure*}

\section{Related Work}
The efficacy of hand gestures as an interaction modality for MR applications on smartphones and HMDs has been extensively explored in the past~\cite{hurst2013gesture}. Marker-based finger tracking~\cite{buchmann2004fingartips} has been established as an effective way of directly manipulating objects in MR applications. However, most of the work has been based either on skin colour or on hand-crafted features for hand segmentation and interest point detection which is followed by optical flow for tracking.

Accurate hand segmentation is very important in all First-Person View (FPV) gesture recognition applications. In early attempts, it was observed that the $YC_bC_r$ colour space allows better clustering of hand skin pixel data~\cite{morerio2013hand}. In~\cite{baraldi2014gesture}, super-pixels with several features are extracted using SLIC algorithm for computing hand segmentation masks. Li et al.~\cite{li2013pixel} observed the response of \textit{Gabor} filters to examine local appearance features in skin colour regions. Most of these approaches are faced by the following challenges: \textit{(i)} movement of the hand relative to the HMD renders the hand blurry, which makes it difficult to detect and track it, thereby impeding classification accuracy. \textit{(ii)} sudden changes in illumination conditions and the presence of skin-like colours and texture in the background causes algorithms with skin feature dependency to fail. 

To this end, we look at utilising the current state-of-the-art object detection architectures like MobileNetV2~\cite{sandler2018mobilenetv2}, YOLOv2~\cite{redmon2016yolo9000} and Faster R-CNN~\cite{ren2015faster} for hand detection. Recently, a Faster R-CNN based hand detector was proposed~\cite{huang2016pointing}. They used a cascaded CNN approach for jointly detecting the hand and the key point using \textit{HSV} colour space information. A dual-target CNN takes input from the Faster R-CNN and localises the index fingertip and the finger-joint.

There are many classification approaches proposed in the context of hand gesture recognition. DTW and HMM based classifiers~\cite{liu2014comparison} have been used with stereo camera setup to recognise third-person view gestures. Support Vector Machines have also been explored for hand gesture recognition via bag-of-features~\cite{dardas2011real}. All such classifiers work well given a small set of sufficiently distinct gestures but fail to extract discriminative features as one scales up to large datasets containing gestures with high inter-class similarly.

Several works~\cite{tompson2014real} use depth information from sensors such as the Microsoft Kinect that restricts its applications in head mounted devices. Moreover, most depth sensors perform poorly in the presence of strong specular reflection, direct sunlight, incandescent light and outdoor environments due to the presence of infrared radiation~\cite{fankhauser2015kinect}. On-device gesture recognition is especially challenging due to the limited sensors present on a smartphone. A recent work~\cite{fink2018gesture} uses a thermographic camera for detecting the infrared radiation from the hand. This method needs additional hardware in the form of an infrared transducer. To the best of our knowledge, ours is the first attempt to make an on-device gesture classification framework for mobile devices.

\section{Proposed Framework}
There has been an increased emphasis on developing \textit{end-to-end} networks that learn to model a number of sub-problems implicitly while training. While this has several advantages in learning joint tasks like object-detection followed by classification, it usually relies on the presence of a large amount of labelled data which has discriminative features useful for each of the sub-problems. For example, consider a scenario where one had large volumes of labelled data for object detection but only a small volume for object detection with classification. In such a scenario, an \textit{end-to-end} model would be restricted to train on the small volume of data available for both tasks, whereas a model that is separately trained on a large detection dataset in conjunction with a classifier is likely to do much better owing to superior detection performance.

Hence, we propose an ensemble of architectures capable of recognising a variety of hand gestures for frugal AR wearable devices with a monocular RGB camera input that requires only a limited amount of labelled classification data. The \emph{GestARLite} architecture is capable of classifying fingertip motion patterns into different hand gestures. Figure~\ref{fig:framework} shows the building blocks of our proposed pipeline: \textit{(i)} \textit{MobileNetV2}~\cite{sandler2018mobilenetv2} takes a single RGB image as an input and outputs a hand candidate bounding box. The input images are first down-scaled to $640\times480$ resolution to reduce processing time without compromising on the quality of image features. \textit{(ii)} The detected hand candidates are then fed to a \textit{fingertip regressor} as depicted in Figure~\ref{fig:regressionframework} which outputs the spatial location of the fingertip. \textit{(iii)} The collection of these is then fed it to the \textit{Bi-LSTM} network for classifying the motion pattern into different gestures. 

The Datasets section discusses the gesture patterns of our \textit{EgoGestAR} dataset used for classification. As there is no static reference for the camera, small errors introduced due to the relative motion of the head with respect to the hand can be rectified by the Bi-LSTM network used for classification. We observe that the Bi-LSTM network is robust to unexpected impulses arising in gesture pattern due to false detections.

\subsection{Hand Detection}
We compared state-of-the-art object detection approaches \textit{(i)} \textit{Faster R-CNN}~\cite{ren2015faster}, \textit{(ii)} \textit{YOLOv2}~\cite{redmon2016yolo9000} and \textit{(iii)} MobileNetV2~\cite{sandler2018mobilenetv2} to detect the specific \textit{pointing hand} pose. 

\textbf{MobileNetV2}~\cite{sandler2018mobilenetv2} is a streamlined architecture that uses depth-wise separable convolutions to build light weight deep neural networks. The depth-wise separable convolution factorises a standard convolution into a depth-wise convolution and a $1\times1$ convolution also called a point-wise convolution thereby reducing the number of parameters in the network. It builds upon the ideas from MobileNetV1~\cite{howard2017mobilenets}, however, it incorporates two new features to the architecture: \textit{(i)} linear bottlenecks between the layers, and \textit{(ii)} skip connections between the bottlenecks. The bottlenecks encode the model's intermediate inputs and outputs while the inner layer encapsulates the model's ability to transform from lower-level concepts such as pixels to higher level descriptors such as image categories. Skip connections, similar to the traditional residual connections, enable faster training without any loss in accuracy. 



In our experiments to detect the hand candidate in RGB input images obtained from wearable devices, we evaluate the MobileNetV2 feature extractor with SSDLite~\cite{sandler2018mobilenetv2} object detection module. The Experiments and Results section highlights the results in comparison with YOLOv2~\cite{redmon2016yolo9000} and Faster R-CNN~\cite{ren2015faster} with a pre-trained VGG-16 model~\cite{simonyan2014very} consisting of 13 shared convolutional layers along with other compact models such as ZF~\cite{zeiler2014visualizing} and VGG1024~\cite{chatfield2014return} by modifying the last fully connected layer to detect hand class (pointing gesture pose).

\subsection{Fingertip Localisation}
We propose a \emph{fingertip regressor} based on a CNN architecture to localise the $(x, y)$ coordinates of the fingertip. The hand candidate detection (pointing gesture pose), discussed in the previous section, triggers the regression CNN for fingertip localisation. The hand candidate bounding box is first cropped and resized to $99\times99$ resolution before feeding it to the network described in Figure~\ref{fig:regressionframework}. 

The proposed architecture consists of two convolutional blocks each with three convolutional layers followed by a max-pooling layer. Finally, we use three fully connected layers to regress over the two coordinate values of fingertip point at the last layer. Since the aim is to determine continuous valued outputs corresponding to fingertip positions, we use Mean Squared Error (MSE) measure to compute loss at the last fully connected layer. The model is trained for robust localisation, and we compare our model with the architecture proposed by Huang et al. ~\cite{huang2016pointing}. They localise both fingertip and finger-joint while we regress fingertip alone.

\subsection{Gesture Classification}
The fingertip localisation network outputs the spatial locations of the fingertip $(x, y)$, which are then fed as an input to our gesture classification network. To reduce computational cost, we just input the $(x,y)$ coordinate instead of the entire frame to the network thereby helping achieve real-time performance. Motivated by the effectiveness of LSTMs~\cite{hochreiter1997long} in learning long-term dependencies of sequential data~\cite{tsironi2016gesture}, we employ a Bi-LSTM~\cite{graves2005framewise} architecture for the classification of gestures. We found that Bi-LSTMs perform better than LSTMs for the particular classification task since they process the sequence in both forward and reverse direction. The usage of LSTMs inherently means that the entire framework is also adaptable to videos and live feeds with variable length frame sequences. This is particularly important as the length of gestures depends on the user performing it and on the performance of the preceding two networks.

Hegde et al.~\cite{hegde2016gestar} conducted a feasibility study for ranking the available modes of interaction for frugal Google Cardboard set-up and reported that the frequent usage of magnetic trigger and conductive lever leads to wear and tear of the device and it scored poorly on usability. Hence, we propose an automatic, implicit trigger to signify the starting and ending of a user input sequence. In the event of a positive pointing-finger hand detection on five consecutive frames, the framework is triggered to start recording the spatial location of the fingertip. Similarly, the absence of any hand detections on five consecutive frames denotes the end of a gesture. The recorded sequence is fed as an input to the \textit{Bi-LSTM} layer consisting of $30$ units. The \textit{forward} and \textit{backward} activations are multiplied before being passed on to the next flattening layer that makes the data one-dimensional. It is then followed by a fully connected layer with $10$ output scores that correspond to each of the $10$ gestures. Since the task is to classify $10$ gesture classes, we use a \textit{softmax} activation function that interprets the output scores as unnormalised log probabilities and squashes the output scores to be between $0$ and $1$ using the following equation:

\begin{equation}
\sigma \left ( s \right )_j = \frac{e^{s_j}}{\sum_{k=1}^{K}e^{s_k}}
\end{equation}
\\
where $K$ denotes number of classes, $s$ is a $K$x$1$ vector of scores, an input to \textit{softmax} function, and $j$ is an index varying from $1$ to $K$. $\sigma \left ( s \right )$ is $K$x$1$ output vector denoting the posterior probabilities associated with each gesture. The \textit{cross-entropy} loss has been used in training to update the model in network back-propagation.

\begin{figure}[t]
    \centering
    \includegraphics[width=0.8\linewidth]{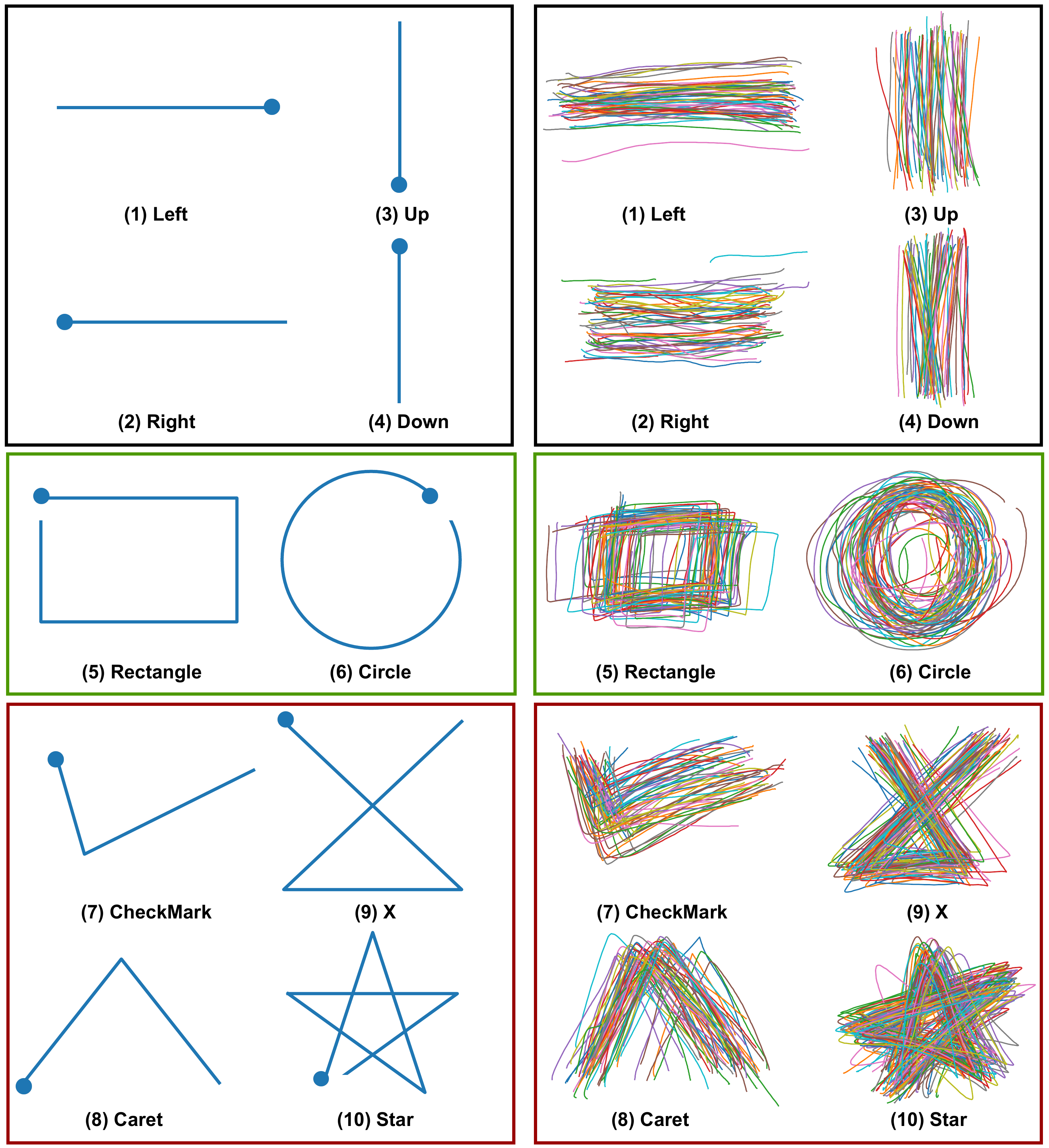}
    \caption{EgoGestAR dataset: The left column shows the standard gesture sequences shown to the users before the data collection and the right column depicts the variation in the collected data samples. The highlighted point in each gesture indicates the starting position of the gesture.}
    \label{fig:dataset}
\end{figure}

\section{Datasets}
\subsection{Hand Dataset}
We use the SCUT-Ego-Finger Dataset~\cite{huang2015deepfinger} for training the hand detection and the fingertip localisation modules. The dataset includes $93729$ frames of pointing hand gesture including hand candidate bounding boxes and index finger key-point coordinates. 


\subsection{EgoGestAR Dataset}
A major factor that has hampered the advent of deep learning in the task of recognising temporal hand gestures is lack of available large-scale datasets to train neural networks on. To our knowledge, there exists no dataset, other than the ones mentioned in this paper, that provides annotated data of a wide range of intuitive temporal hand gestures.

Hence, to train and evaluate the proposed gesture classification network, we propose \textbf{\textit{EgoGestAR}}: an egocentric-vision gesture dataset for AR/MR wearables. The dataset includes 10 gesture patterns. To introduce variability in the data, the dataset has been collected with the help of $50$ subjects chosen at random from our research lab with ages spanning from $21$ to $50$. The average age of the subjects was $27.8$ years. The dataset consists of $2500$ gesture patterns where each subject recorded $5$ samples of each gesture. The gestures were recorded by mounting a tablet PC (model HP10EEG1) to a wall. The patterns drawn by the user's index finger on a touch interface application with position sensing region was stored. The data was captured at a resolution of $640\times480$. Figure~\ref{fig:dataset} describes the standard input sequences shown to the users before data collection and a sample subset of gestures from the dataset showing the variability introduced by the subjects. 
These gestures primarily divided into 3 categories for effective utilisation in our context of data visualisation in MR applications:

\begin{itemize}
    \item[(i)] 4 swipe gesture patterns (\textit{Up, Down, Left, and Right}) for navigating through graph visualisations/lists.
    \item[(ii)] 2 gesture patterns (\textit{Rectangle and Circle}) for RoI highlighting in user's FoV and for zoom-in and zoom-out operations.
    \item[(iii)] 4 gesture patterns (\textit{CheckMark: Yes, Caret: No, X: Delete, Star: Bookmark}) for answering contextual questions while interacting with applications such as industrial inspection~\cite{7836501}.
\end{itemize} 

Further, to test the entire framework, $240$ videos were recorded by a random subset of the aforementioned subjects performing each gesture $22$ times. Additional $20$ videos of random hand movements were also recorded. The videos were recorded using a OnePlus X Android device mounted on a Google Cardboard. High quality videos are captured at a resolution of $640$x$480$, and at $30$ \textit{FPS}. We have published the dataset online~\footnote{\url{https://varunj.github.io/gestarlite}} for the benefit of the research community.

\section{Experiments and Results}
Since the framework comprises of three networks, we evaluate the performance of each of the networks individually to arrive at the best combination of networks for our proposed application. We use an 8 core Intel\textsuperscript{\textregistered} Core\textsuperscript{\texttrademark} i7-6820HQ CPU, 32GB memory and an Nvidia\textsuperscript{\textregistered} Quadro M5000M GPU machine for experiments. A Snapdragon\textsuperscript{\textregistered} 845 chip-set smartphone was used which was interfaced with the server (wherever needed: to evaluate our method that runs on device) using a local network hosted on a Linksys EA6350 802.11ac compatible wireless router.

For all the experiments pertaining to hand detection and fingertip localisation, we use the hand dataset~\cite{huang2015deepfinger}. Out of the $24$ subjects present in the dataset, we choose $17$ subjects' data for training with a validation split of $70$:$30$, and $7$ subjects' data ($24,155$ images) for testing the networks.

\begin{figure}[ht]
    \centering
    \includegraphics[width=0.9\linewidth]{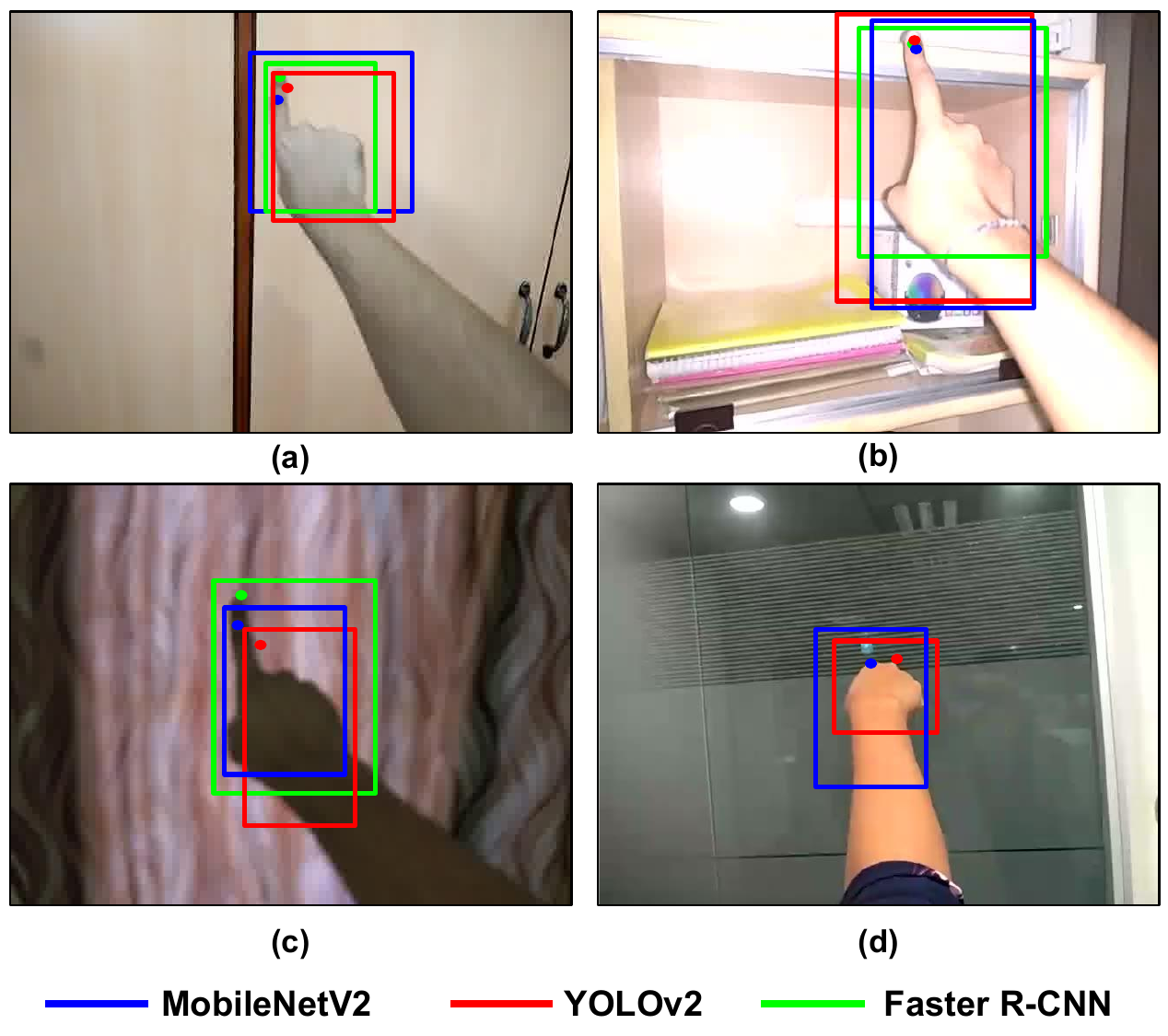}
    \caption{Hand candidate detection and fingertip localisation on sample images using MobileNetV2, YOLOv2 and Faster R-CNN with ZF networks in several challenging conditions. \textit{(a)} indoor light with skin-like colours in background, \textit{(b)} specular reflection, \textit{(c)} low light with blurry hand, and, \textit{(d)} hand in the absence of the pointing gesture. MobileNetV2 localises the hand more accurately as compared to YOLOv2 by giving tighter bounds on hands. Faster R-CNN is the most accurate, but does not work on-device.}
    \label{fig:detect_compare}
\end{figure}

\begin{table}
    \begin{center}
    \resizebox{\linewidth}{!} {
        \begin{tabular}{|l|c|c|c|c|c|}
        \hline
        \makecell{\textbf{Model}} & \makecell{On \\ Device} & \makecell{mAP \\ IoU=0.5} & \makecell{mAP \\ IoU=0.7} & \makecell{rate\\(FPS)} & \makecell{Model\\Size}\\
        \hline
        \makecell{F-RCNN\\VGG16} & \xmark & 98.1 & 86.9 & 3 & 546 MB\\
        \hline
        \makecell{F-RCNN\\VGG1024} & \xmark & 96.8 & 86.7 & 10 & 350 MB\\
        \hline
        \makecell{F-RCNN\\ZF} & \xmark & 97.3 & 89.2 & 12 & 236 MB\\
        \hline
        \makecell{YOLOv2} & \checkmark & 93.9 & 78.2 & 2 & 202 MB\\
        \hline
        \makecell{\textbf{MobileNetV2}} & \checkmark & \textbf{89.1} & \textbf{85.3} & \textbf{9} & \textbf{12 MB}\\
        \hline
        \end{tabular}
    }
    \end{center}
    \caption{Performance of various methods on the \textit{SCUT-Ego-Finger} dataset for hand detection. mAP score, frame-rate and the model size are reported with the variation in IoU.}
    \label{table:f1_measure}
\end{table}

\subsection{Hand Detection}
Table~\ref{table:f1_measure} reports percentage of mean Absolute Precision (mAP) and frame rate for hand candidate detection. Even though MobileNetV2~\cite{sandler2018mobilenetv2} achieved higher frame-rate compared to others, it produced high false positives hence resulted in poor classification performance. It is observed that YOLOv2 can also run on-device although it outputs fewer frames as compared to MobileNetV2. At an Intersection over Union (IoU) of 0.5, YOLOv2 achieves $93.9\%$ mAP on SCUT-Ego-Finger hand dataset whereas MobileNetV2 achieves $89.1\%$ mAP. However, we observe that YOLOv2 performs poorly when compared to MobileNetV2 in localising the hand candidate at higher IoU that is required for including the fingertip. Figure~\ref{fig:detect_compare} shows results of the detectors in different conditions such as poor illumination, blurry rendering, indoor and outdoor environments. We notice that even though both the detectors are unlikely to predict false positives in the background, YOLOv2 makes more localisation errors proving MobileNetV2 to be a better fit for our use-case. 

It is worth noticing that the model size for MobileNetV2 is significantly less than the rest of the models. It enables us to port the model on mobile device and removes the framework's dependence on a remote server. This helps reduce latency introduced by the network and can enable a enable wider reach of frugal devices for MR applications.

\begin{figure}[ht]
    \centering
    \includegraphics[width=1.0\linewidth]{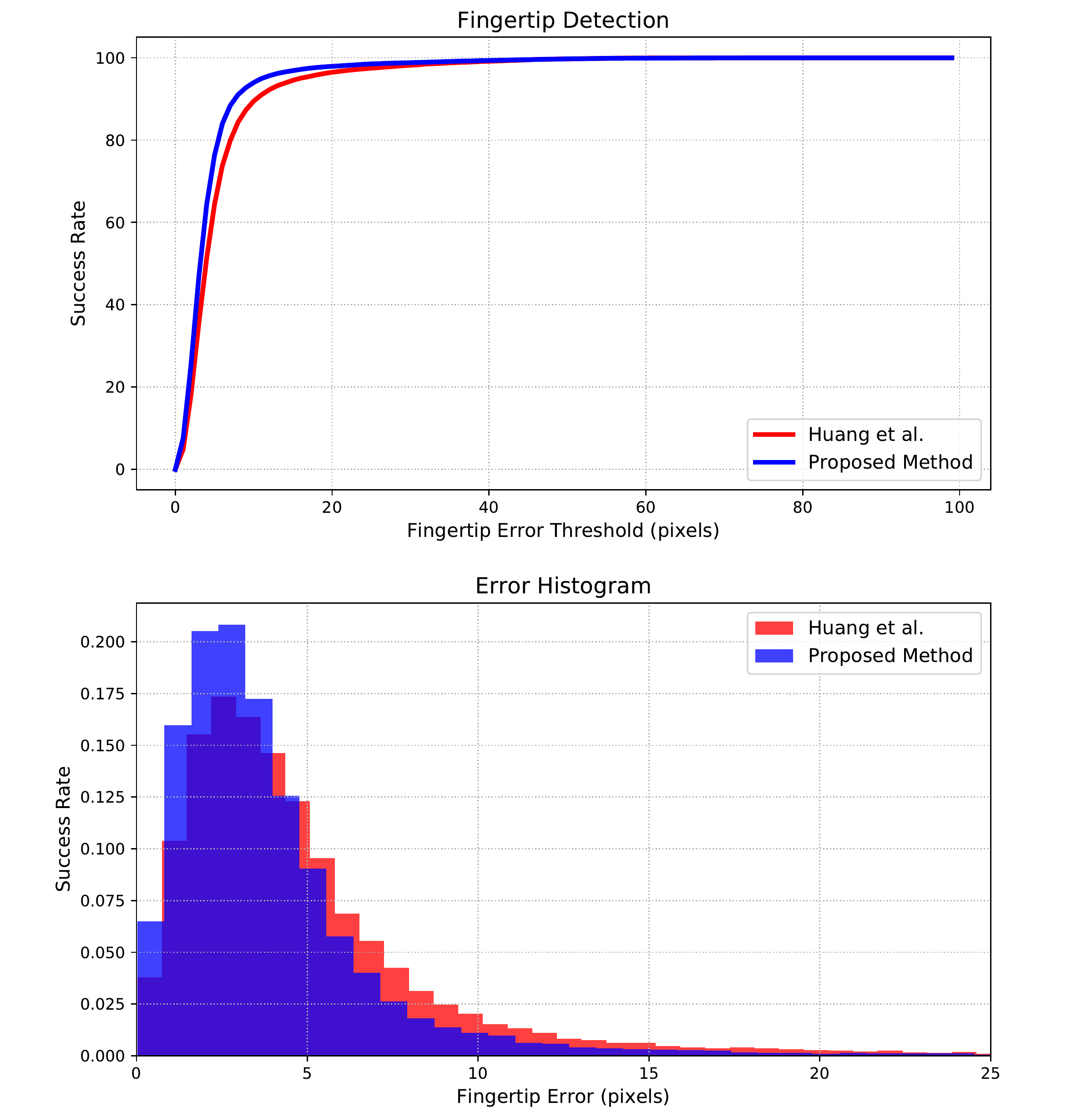}
    \caption{Comparison of our proposed finger localisation model with Huang et al.~\cite{huang2016pointing}. \textit{(above)} our model achieves a higher success rate at any given error threshold. \textit{(below)} The fraction of images with low localisation error is higher for our proposed method.}
    \label{fig:reg_result}
\end{figure}

\subsection{Fingertip Localisation}
We evaluate the model employed for fingertip localisation on the test set of $24,155$ images. The $2\times1$ continuous-valued output corresponding to finger coordinate estimated at the last layer are compared against ground truth values to compute rate of success with changing thresholds on the error (in pixels) and the resultant plot when compared to the network proposed by Huang et al.~\cite{huang2016pointing} is shown in Figure~\ref{fig:reg_result}. \textit{Adam} optimiser with a learning rate of $0.001$ has been used. The model achieves $89.06\%$ accuracy with an error tolerance of $10$ pixels on an input image of $99\times99$ resolution. The mean absolute error is found to be $2.72$ pixels for our approach and is $3.59$ pixels for the network proposed in~\cite{huang2016pointing}.

\begin{table}
    \begin{center}
    \begin{tabular}{|l|c|c|c|c|}
    \hline
    \makecell{Method} & \makecell{Precision} & \makecell{Recall} & \makecell{$F_1$ Score} \\
    \hline
    \makecell{DTW~\shortcite{liu2014comparison}} & 0.741 & 0.76 & 0.734\\
    \hline
    \makecell{SVM~\shortcite{fan2008liblinear}} & 0.860 & 0.842 & 0.851\\
    \hline
    \makecell{LSTM~\shortcite{hochreiter1997long}} & 0.975 & 0.920 & 0.947\\
    \hline
    \makecell{\textbf{Bi-LSTM}~\shortcite{graves2005framewise}} & \textbf{0.956} & \textbf{0.940} & \textbf{0.948}\\
    \hline
    \end{tabular}
    \end{center}
    \caption{Performance of different classification methods on our EgoGestAR dataset. Average of precision and recall values for all classes is computed to get a single number.}
    \label{table:lstm_dtw}
\end{table}

\subsection{Gesture Classification}
We use our \textit{EgoGestAR} dataset for training and testing of the gesture classification network. We also tried classification with an LSTM network in the same training and testing setting as the Bi-LSTM. During training, $2000$ gesture patterns of the training set were used. A total of $8,230$ parameters of the network are trained with a batch size of $64$ and validation split of $80:20$. \textit{Adam} optimiser with learning rate of $0.001$ has been used. The networks are trained for $900$ epochs and achieved validation accuracy of $95.17\%$ and $96.5\%$ for LSTM and Bi-LSTM respectively. LSTM and Bi-LSTM achieve classification accuracy of $92.5\%$ and $94.3\%$ respectively, outperforming the traditional approaches that are being used for similar classification tasks. We present comparison of the proposed LSTM and Bi-LSTM approach with DTW~\cite{liu2014comparison} and SVM~\cite{fan2008liblinear} classification in Table~\ref{table:lstm_dtw}. Additionally, we observe that the performance of traditional methods like DTW and SVM deteriorate significantly in the absence of sufficient data-points. Hence, they rely on complex interpolation techniques to give consistent results.

\begin{figure}[ht]
    \centering
    \includegraphics[width=0.9\linewidth]{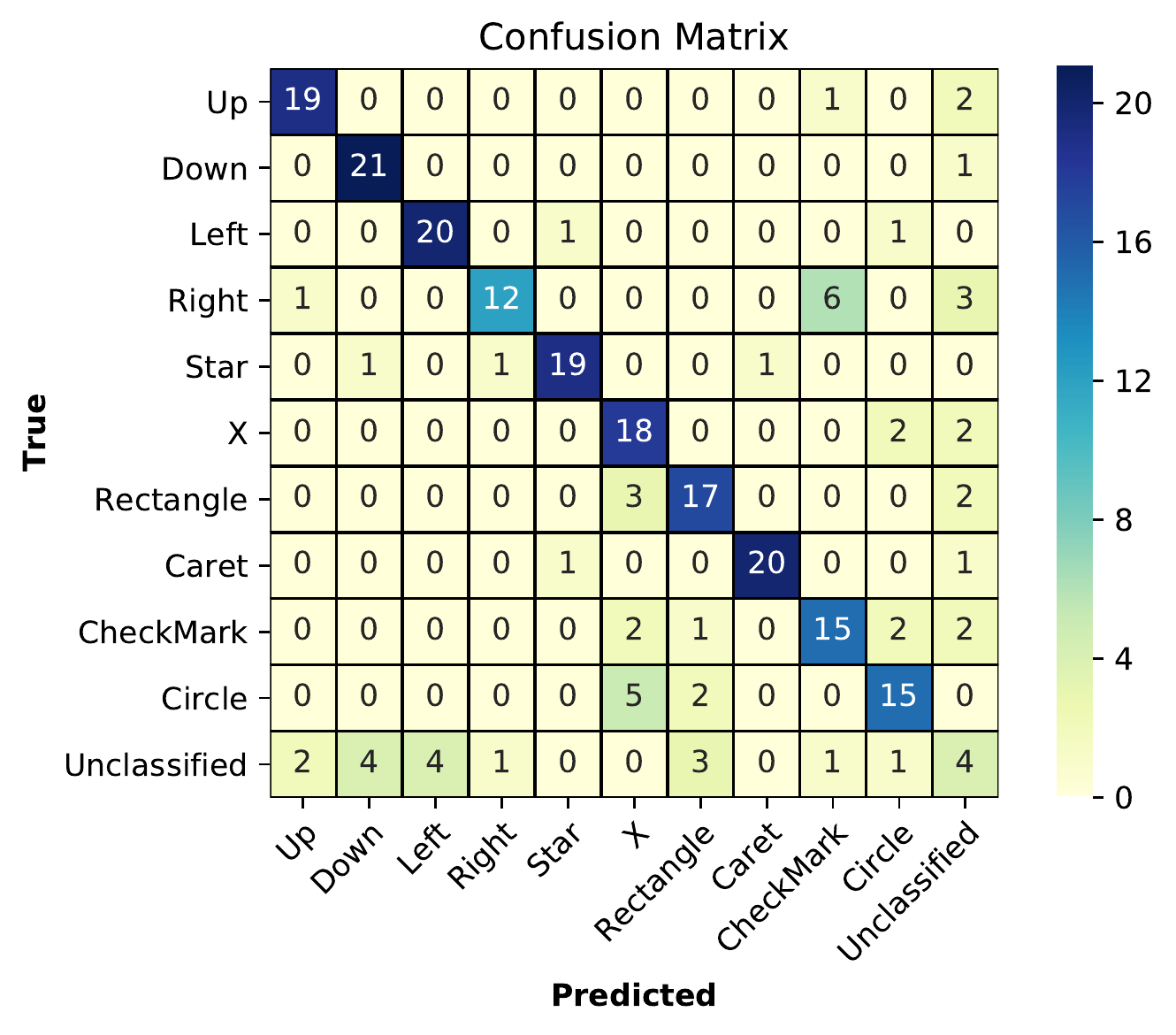}
    \caption{The overall performance of our proposed framework on 240 egocentric videos captured using a smartphone based \textit{Google Cardboard} head-mounted device. The gesture is detected when the predicted probability is more than $0.85$. Accuracy of our proposed framework is $0.8$ (excluding the unclassified class).}
    \label{fig:framework_result}
\end{figure}

\subsection{Framework Evaluation}
Since the proposed approach is a series of different networks, the overall classification accuracy in real-time will vary depending on the performance of each network used in the pipeline. Therefore, we evaluate the entire framework using 240 egocentric videos captured with a smartphone based Google Cardboard head-mounted device. The MobileNetV2 model was used in our experiments as it achieved the best trade-off between accuracy and performance. Since the model can work independently on a smartphone using the TF-Lite engine, it removes the framework's dependence on a remote server and a quality network connection.

The framework achieved an overall accuracy of $80.00\%$ on a dataset of 240 egocentric videos captured in FPV is as shown in Figure~\ref{fig:framework_result}. The MobileNetV2 network works at $9$ \textit{FPS} on $640\times480$ resolution videos, and the \textit{fingertip regressor} is capable of delivering frame rates of up-to $166$ \textit{FPS} working at a resolution of $99\times99$. The gesture classification network is capable of processing a given stream of data in less than $100$\textit{ms}. As a result, the average response time of the proposed framework is found to be $0.12$\textit{s} on a smartphone powered by a Snapdragon\textsuperscript{\textregistered} 845 chip-set. The entire model has a very small memory footprint of $16.3$ \textit{MB}.

\begin{table}
    \begin{center}
    \begin{tabular}{|l|c|c|c|c|}
    \hline
    \makecell{Method} & \makecell{Accuracy} & \makecell{Time taken} & \makecell{On Device} \\
    \hline
    \makecell{Tsironi et al.} & 32.27 & 0.76 & \xmark \\
    \hline
    \makecell{VGG16+LSTM} & 58.18 & 0.69 & \xmark \\
    \hline
    \makecell{C3D} & 66.36 & 1.19 & \xmark \\
    \hline
    \makecell{\textbf{GestARLite}} & \textbf{80.00} & \textbf{0.12} & \checkmark \\
    \hline
    \end{tabular}
    \end{center}
    \caption{Analysis of gesture recognition accuracy and latency of various models against the proposed framework. Our framework GestARLite works on-device and effectively has the highest accuracy and the least response time.}
    \label{table:comparison}
\end{table}

\section{Discussion and Comparison}
Further analysis of the results shows that the \textit{CheckMark} gesture is slightly correlated with the \textit{Right} gesture since they both involve an arc that goes from left to right. Hence we observe a drop in the classification accuracy due to the inherent subjectivity of how a user draws a given gesture. Further, there are cases when the user's hand goes out of the frame which can lead to a gesture not getting classified. Our framework is currently limited to a single finger in the users' FoV and the accuracy drops if multiple fingers are present at roughly the same distance. However, we observe that the framework performs equally well in cases where the user is wearing nail paint or has minor finger injuries. In such cases, the \textit{fingertip regressor} outputs the point just below the nail and tracks it. Further, it is also robust to different hand colours and sizes. The EgoGestAR dataset can potentially be extended to a number of pointing gestures as per the requirement of the MR application since the framework can accommodate a number of sufficiently distinct gestures.

We also make an important observation regarding our approach for including dedicated modules in our pipeline in contrast to an \textit{end-to-end} framework. We compared our modular pipeline against several \textit{end-to-end} trained gesture classification methods Table~\ref{table:comparison}. Tsironi et al.\cite{tsironi2016gesture} proposed a network that works with differential image input to convolutional LSTMs to capture the body parts' motion involved in the gestures performed in second-person view. Even after fine-tuning the model on our egocentric video dataset, it produced an accuracy of only $32.14\%$ as our data involved a dynamic background and no static reference to the camera. 

The VGG16+LSTM network~\cite{donahue2015long} uses 2D CNNs to extract features from each frame. These frame-wise features are then encoded as a temporally deep video descriptor which is fed to an LSTM network for classification. Similarly, a 3D CNNs approach~\cite{tran2015learning} uses 3D CNNs to extract features directly from video clips. Table~\ref{table:comparison} shows that both of these methods do not perform well. A plausible intuitive reason for this is that the network might be learning noisy and bad features while training. 

Attention based video classification~\cite{sharma2015action} also performed poorly owing to the high inter-class similarity. Since we require features from only a small portion of the entire frame, that is, the fingertip, such attention models appear redundant since we already know the fingertip location.

\section{Conclusion}
We have presented \emph{GestARLite} to enable mass market reach of gestural interfaces that works in a resource constrained environment like on a smartphone or a video see-through HMD. \emph{GestARLite} works in real-time and on-device achieving an accuracy of $80.00\%$ with a model size of $16.3$ \textit{MB}. Our approach is marker-less and uses only RGB data from a single smartphone camera. The framework is tested on GPUs and Android devices with a Snapdragon\textsuperscript{\textregistered} 845 chip-set. To the best of our knowledge, our proposed work is the first of its kind, deep learning based attempt to build on-device gestural interface for frugal head-mounted devices without the built-in depth sensors and having no need of network connectivity.

In the future, we intend to use  the \emph{GestARLite} architecture to come up applications of gesture recognition in second person view on Android phones exclusively for the visually challenged. Further, we aim to develop applications for huge display screens and boardrooms as an attempt to replace the mouse pointer with fingertip for controlling presentations.

\bibliography{aaai.bib}
\bibliographystyle{aaai}
\end{document}